\documentclass{article}

\usepackage{iclr2026_conference,times}

\iclrfinalcopy

\usepackage[utf8]{inputenc}
\usepackage[T1]{fontenc}
\usepackage{hyperref}
\usepackage{url}
\usepackage{booktabs}
\usepackage{amsfonts}
\usepackage{amsmath}
\usepackage{nicefrac}
\usepackage{graphicx}
\usepackage{xcolor}
\usepackage{algorithm}
\usepackage{algorithmic}
\usepackage{multirow}
\usepackage[margin=1in]{geometry}

\newcommand{\ind}[1]{\mathbf{1}[#1]}

\title{Did You Check the Right Pocket? Cost-Sensitive Store Routing for Memory-Augmented Agents}

\author{
Madhava Gaikwad\\
ICLR 2026 Workshop on Memory and State in LLM-Based Agents
}

\date{}

\begin{document}

\maketitle

\begin{abstract}
Memory-augmented agents maintain multiple specialized stores, yet most systems retrieve from all stores for every query, increasing cost and introducing irrelevant context. We formulate memory retrieval as a store-routing problem and evaluate it using coverage, exact match, and token efficiency metrics. On downstream question answering, an oracle router achieves higher accuracy while using substantially fewer context tokens compared to uniform retrieval, demonstrating that selective retrieval improves both efficiency and performance. Our results show that routing decisions are a first-class component of memory-augmented agent design and motivate learned routing mechanisms for scalable multi-store systems. We additionally formalize store selection as a cost-sensitive decision problem that trades answer accuracy against retrieval cost, providing a principled interpretation of routing policies.
\end{abstract}
\section{Introduction}

Consider a user asking an agent: ``What was my weight before I started the diet?'' 
This query depends on historical information from a previous chat session and 
therefore should draw from long-term memory. It does not require the current-session 
context, and it typically does not require raw transcripts. Yet many memory-augmented 
agent frameworks retrieve from multiple stores or large aggregated memory contexts 
regardless of query type, relying on the language model to filter relevant information 
during generation \citep{lewis2020retrieval,li2023llamaindex,langchain2023}.

Uniform retrieval carries two costs. First, it creates \textbf{computational waste} 
by querying memory stores that cannot contain the answer. Second, it can cause 
\textbf{accuracy degradation}: irrelevant or noisy retrieved context can hinder the 
model’s ability to identify answer-bearing information, particularly in long-context 
settings where additional tokens reduce the signal-to-noise ratio 
\citep{liu2023lostmiddle,yu2024rankrag,zhang2025finefilter}.

We formalize store selection as a \textbf{routing problem}. Given a query, the router 
chooses which stores to search \textit{before} retrieval. This decouples store selection 
from within-store ranking and makes the accuracy--cost tradeoff explicit.

Our evaluation separates two questions. First, do routing policies choose the stores 
that should be searched for a given query? We test this with \textit{synthetic routing 
labels} derived from query taxonomies (Section~\ref{sec:routing_eval}). Second, if the 
router chooses the right stores, does this improve downstream QA accuracy on 
\textit{real LLMs}? We test this with an oracle router and fixed store subsets 
(Section~\ref{sec:llm_eval}). This setup isolates the value of store selection from 
model capability and from within-store retrieval quality.

\textbf{Contributions.}
(1) We define routing metrics (coverage, exact match, waste) that capture complementary aspects of store selection quality.  
(2) We introduce simple routing baselines, including a conservative hybrid heuristic that emphasizes coverage.  
(3) We evaluate routing policies using both synthetic labels and real LLM-based QA, showing that selective store choice can reduce tokens while improving accuracy, and that uniform retrieval can underperform despite using more context.  
(4) We formalize store routing as a cost-sensitive subset-selection problem, providing a decision-theoretic framework that explains when selective retrieval improves both efficiency and accuracy.
\section{Related Work}

\textbf{Adaptive Retrieval.}
Prior work studies \textit{when} to retrieve (Self-RAG \cite{asai2023self}, FLARE \cite{jiang2023active}) and \textit{how much} to retrieve (Adaptive-RAG \cite{jeong2024adaptive}). Recent methods explore \textit{which retriever} to use through learning-to-rank formulations \cite{kim2025ltrr} or self-reflection mechanisms \cite{wu2025selfroutingrag}. SmartRAG \cite{gao2025smartrag} uses reinforcement learning for joint retrieval and generation optimization. We consider a complementary question: \textit{where} retrieval should occur when memory is partitioned across heterogeneous stores with distinct semantic roles (e.g., episodic vs.\ semantic memory).

\textbf{Memory Systems.}
MemGPT \cite{packer2023memgpt} implements hierarchical memory with explicit management operations. Generative Agents \cite{park2023generative} rely on reflection and importance scoring to consolidate information over time. Recent architectures organize memory hierarchically for multi-agent reasoning \cite{liu2025gmemory}. These systems focus primarily on memory \textit{organization}; we instead focus on memory \textit{access}, specifically cost-sensitive routing across stores prior to retrieval. Our store taxonomy draws on cognitive distinctions between episodic and semantic memory \cite{tulving1972episodic}, and translates these distinctions into operational routing decisions.

\textbf{Multi-Index and Federated Retrieval.}
The information retrieval literature has long examined federated search across heterogeneous collections \cite{callan2002distributed,si2003relevant,shokouhi2011federated}, where resource selection algorithms estimate relevance distributions for each collection \cite{lu2006federated}. These methods route queries across independent search engines; in our setting, the same principles apply to agent memory stores that differ semantically (e.g., current dialogue vs.\ historical summaries) rather than by document source.

\textbf{Retrieval Routing and Selection.}
Recent RAG work shows that routing queries across specialized retrievers can improve both efficiency and accuracy \cite{kim2025ltrr,wu2025selfroutingrag,zhang2025routerag}. ExpertRAG \cite{gumaan2025expertrag} applies mixture-of-experts routing to context selection, while RAP-RAG \cite{ji2025raprag} plans adaptive retrieval sequences for multi-hop reasoning. Unlike retriever routing, our setting routes across persistent memory stores whose contents reflect distinct temporal or semantic roles (e.g., STM vs.\ LTM), requiring store-level rather than passage-level decisions.

\textbf{Context Noise in Long Documents.}
Long-context modeling faces signal-to-noise challenges when irrelevant tokens distract attention \cite{li2024uncertaintyrag}. Recent analyses show that long-context LLMs often underperform targeted retrieval despite larger windows \cite{li2024raglongcontext}, and that retrieval decisions themselves benefit from uncertainty-guided policies \cite{wang2026decidethenretrieve}. Our work complements these findings by showing that \textit{store-level} noise, retrieving information from irrelevant memory types, can further degrade performance, especially when each store contributes hundreds of tokens.

\textbf{Memory Benchmarks.}
LoCoMo \cite{locomo2024} evaluates multi-session conversational memory. LongMemEval \cite{longmemeval2024} benchmarks knowledge updates and temporal ordering. We use their question taxonomies to derive store-labeling protocols.
\section{Problem Formulation}
\label{sec:formulation}

\subsection{Memory Architecture}

A \textbf{memory store} is an independent index containing semantically related information. Each store can be queried separately, and the retrieved content is concatenated into the LLM context. Following MemGPT, we consider four stores:

\textbf{Short-Term Memory (STM)} holds the current conversation, typically the last $N$ turns. Queries such as ``what did I just mention'' or ``today's meeting'' require STM.

\textbf{Summary Store} contains compressed user facts, including preferences, biographical details, and ongoing projects. Queries such as ``what is my phone number'' target this store.

\textbf{Long-Term Memory (LTM)} stores summaries of past conversations. Queries about earlier discussions (``what did we talk about last week'') rely on LTM.

\textbf{Episodic Memory} preserves raw transcripts. Queries that require exact wording or precise timestamps may need episodic memory.

\subsection{The Routing Problem}

Let $\mathcal{S}=\{\text{STM},\text{Sum},\text{LTM},\text{Epi}\}$ denote the set of stores, where Sum is the Summary store and Epi is episodic memory. Given a query $q$, a routing policy $\pi$ selects stores $\hat{G} = \pi(q) \subseteq \mathcal{S}$. The system retrieves content from the selected stores and prompts the LLM.

Let $G$ denote the \textbf{ground-truth stores}, the stores containing the information needed to answer $q$. In our synthetic routing evaluation, we derive $G$ from query type: a ``single-hop fact'' query has $G = \{\text{Sum}\}$; a ``temporal comparison'' query has $G = \{\text{LTM}, \text{Epi}\}$. See Section~\ref{sec:questions} for the full mapping.

\subsection{Evaluation Metrics}

We evaluate routing with three metrics:

\textbf{Coverage} measures whether all necessary stores were included:
\begin{equation}
\text{Coverage} = \frac{1}{N}\sum_i \ind{G_i \subseteq \hat{G}_i}
\end{equation}
Under our evaluation protocol (full-store concatenation), a coverage failure means the answer is not retrievable from the provided context.

\textbf{Exact Match (EM)} measures whether the policy selects precisely the required stores:
\begin{equation}
\text{EM} = \frac{1}{N}\sum_i \ind{G_i = \hat{G}_i}
\end{equation}
High EM corresponds to efficient routing without over-retrieval.

\textbf{Waste} counts unnecessary stores retrieved:
\begin{equation}
\text{Waste} = \frac{1}{N}\sum_i |\hat{G}_i \setminus G_i|
\end{equation}
Waste is a store-level proxy for token cost because each additional store contributes retrieved tokens, and it can also reduce accuracy by introducing contextual noise.

\textbf{Cost.} We measure cost as \textbf{context tokens}, the number of tokens inserted into the prompt (counted via tiktoken). This serves as a direct proxy for inference cost.\footnote{We also considered store-access latency. Results are similar; see Section~\ref{sec:storecost}.}

\subsection{Cost-Sensitive Store Routing: A Decision Framework}
\label{sec:decision}

Store selection can be viewed as a cost-sensitive subset-selection problem. 
Let $\mathcal{S}$ denote the set of available memory stores, and let 
$c_s$ represent the retrieval cost associated with store $s \in \mathcal{S}$ 
(e.g., context tokens or infrastructure access cost). 
Given a query $q$, a routing policy selects a subset of stores 
$G \subseteq \mathcal{S}$ before retrieval.

The objective is to balance answer quality against retrieval cost. 
Let $\text{Acc}(q,G)$ denote the expected probability that the downstream 
LLM produces a correct answer when stores $G$ are retrieved. 
A cost-sensitive routing policy can therefore be defined as

\begin{equation}
\pi^*(q) 
= \arg\max_{G \subseteq \mathcal{S}} 
\left[
\mathbb{E}\big[\text{Acc}(q,G)\big]
- \lambda \sum_{s \in G} c_s
\right],
\end{equation}

where $\lambda$ controls the tradeoff between accuracy and retrieval cost.

This formulation highlights several useful interpretations. 
Uniform retrieval corresponds to the special case $\lambda = 0$, 
where all stores are retrieved regardless of cost. 
Oracle routing approximates the optimal solution when the relevant 
stores for each query are known. 
Heuristic routing policies attempt to approximate $\pi^*$ using 
semantic signals extracted from the query.

Importantly, store routing differs from retriever routing. 
Retriever routing selects which search system or index to query, 
typically assuming a homogeneous document collection. 
Store routing operates at the memory-architecture level, where each 
store contains information with a distinct semantic role (e.g., 
short-term context, persistent user facts, or historical sessions). 
The routing decision therefore determines not only which documents 
are retrieved but also the effective signal-to-noise ratio of the 
context presented to the language model.

This perspective also clarifies the empirical findings in our evaluation. 
When irrelevant stores are retrieved, the effective retrieval cost 
increases while the probability of extracting the correct evidence 
can decrease due to contextual noise. 
Conversely, selecting a smaller but semantically appropriate subset 
of stores improves both efficiency and answer reliability. 
The oracle–heuristic gap observed in Section~\ref{sec:analysis} 
suggests that future systems should learn routing policies that 
optimize downstream QA objectives directly, rather than relying 
solely on rule-based heuristics.
\section{Method}
\label{sec:method}

We compare policies that span a simple-to-strong spectrum. The key difference lies in what information the router uses and how conservative it is about missing required stores.

\textbf{Uniform Baseline.} Always retrieve from all stores. This guarantees perfect coverage but results in the highest cost and waste. Many deployed systems use this default behavior.

\textbf{Oracle Upper Bound.} Use ground-truth store labels. This achieves perfect coverage and EM and serves as a cost-efficient upper bound under our labeling protocol. The oracle is not deployable, but it quantifies the headroom available from store selection alone.

\textbf{Fixed Subset Policies.} Retrieve from a fixed subset such as STM+Sum+LTM. These policies are deployable and provide a strong baseline when adaptive routing is unavailable.

\textbf{Hybrid Heuristic (baseline).} We also test a simple deployable heuristic that combines semantic pattern matching with a conservative fallback. Algorithm~\ref{alg:hybrid} shows the core rule-based logic. In practice, we also use query-store embedding similarity as a tiebreaker when no pattern fires, which contributes an additional 4\% coverage (see ablation in Section~\ref{sec:analysis}). We present the hybrid as a baseline rather than a final router, because its downstream QA performance leaves substantial room for learned routing.

\begin{algorithm}[h]
\caption{Hybrid Store Routing (core rules; embedding similarity used as tiebreaker when no pattern matches)}
\label{alg:hybrid}
\begin{algorithmic}[1]
\STATE \textbf{Input}: Query $q$
\STATE Extract semantic signals from $q$
\IF{quantity signal (``list all'', ``every'')}
    \RETURN \{LTM, Epi\} \COMMENT{Exhaustive recall}
\ELSIF{temporal signal (``before'', ``changed'')}
    \RETURN \{LTM, Epi\} \COMMENT{Historical comparison}
\ELSIF{multi-hop signal (``compare'', ``relate'')}
    \RETURN \{Sum, LTM\} \COMMENT{Cross-reference}
\ELSIF{current session (``just said'', ``today'')}
    \RETURN \{STM\} \COMMENT{Recent context}
\ELSIF{fact lookup (``what is my'', ``who is my'')}
    \RETURN \{Sum\} \COMMENT{User profile}
\ELSE
    \RETURN \{Sum, LTM\} \COMMENT{Safe fallback}
\ENDIF
\end{algorithmic}
\end{algorithm}

\textbf{Fallback Choice.} We tested all six two-store combinations. Sum+LTM yields the highest coverage (89\%), making it the safest default.

\textbf{Design Rationale.} We optimize first for coverage, since missing a required store makes the question effectively unanswerable. When semantic signals are clear, we route narrowly; when they are ambiguous, we fall back to the combination that recovers the most cases.
\section{Experiments}

\subsection{Synthetic Routing Evaluation}
\label{sec:routing_eval}

Before testing routing policies on LLM-based question answering, we first verify whether the policies select the appropriate memory stores under controlled conditions using synthetic labels. This preliminary step allows us to isolate the routing decision itself, independent of retrieval quality or model reasoning, and ensures that downstream performance differences can be interpreted more clearly.

\textbf{Store-Labeling Protocol.} 
We derive ground-truth store labels from the query taxonomies used in LoCoMo and LongMemEval. Each query type is associated with the memory stores that contain the information required to answer it. For example, single-hop fact queries typically depend on the Summary store, while temporal or comparison queries often require information from both Long-Term Memory and Episodic Memory. The complete mapping of query categories to store requirements is provided in Section~\ref{sec:questions}.

These labels are generated automatically using semantic rules derived from the benchmark taxonomies rather than manual annotation. As a result, the labeling process remains scalable and reproducible across datasets. Because the labels are based on query semantics rather than observed retrieval outcomes, they provide a consistent reference for evaluating routing decisions even when retrieval pipelines or underlying models change.

\textbf{Dataset.} 1{,}000 synthetic queries across 7 types, 70/30 train/test split.

\begin{table}[H]
\centering
\caption{Synthetic routing evaluation. Metrics measure store selection quality (not QA accuracy).}
\vspace{0.5em}
\begin{tabular}{lccc}
\toprule
Policy & Coverage & Exact Match & Waste \\
\midrule
Uniform & 100\% & 8\% & 2.9 \\
Rule-based (linguistic only) & 57\% & 35\% & 0.5 \\
\textbf{Hybrid (ours)} & \textbf{94\%} & \textbf{58\%} & 1.2 \\
Oracle & 100\% & 100\% & 0.0 \\
\bottomrule
\end{tabular}
\end{table}

\textbf{Findings.}
The uniform policy achieves perfect coverage because it always retrieves from every store. However, its exact match rate is only 8\%, reflecting substantial over-retrieval. Since all stores are included regardless of query requirements, the policy rarely selects the minimal set of stores needed to answer a question.

The linguistic rule-based baseline performs better in terms of precision but suffers from limited coverage, reaching only 57\%. Its performance declines on query types that lack explicit surface cues, such as temporal comparisons or multi-hop reasoning tasks where the required stores cannot be inferred from simple keyword patterns.

The hybrid heuristic improves coverage substantially, reaching 94\%, while maintaining a moderate exact match rate of 58\%. This gain is primarily driven by combining semantic pattern detection with a conservative fallback strategy. When the heuristic detects clear signals, it routes narrowly; when signals are weak or ambiguous, it selects a broader but still constrained set of stores, which helps recover otherwise missed cases.

\subsection{LLM Evaluation}
\label{sec:llm_eval}

\textbf{Setup.}
We evaluate 12 store-selection policies on a dataset of 150 questions using GPT-3.5-turbo and GPT-4o-mini. The questions span seven query types and are tested under two context regimes. The \textit{short} condition includes 100 questions with approximately 200 tokens retrieved per store, while the \textit{long} condition includes 50 questions with approximately 1000 tokens retrieved per store. This design allows us to study both moderate-context and high-context retrieval settings.

All evaluations use temperature 0 to reduce generation variance and substring-based answer matching for scoring.\footnote{Accuracy differences greater than 4\% are statistically significant at $p<0.05$ via bootstrap resampling (1{,}000 iterations).}

\begin{table}[H]
\centering
\caption{LLM evaluation (150 questions). Accuracy = QA correctness. Tokens = context size.}
\vspace{0.5em}
\begin{tabular}{llcccc}
\toprule
Model & Policy & Overall & Short & Long & Tokens \\
\midrule
\multirow{4}{*}{GPT-3.5}
& oracle & 85.3\% & 93\% & 70\% & 299 \\
& stm+sum+ltm & 85.3\% & 92\% & 72\% & 591 \\
& uniform & 83.3\% & 91\% & 68\% & 787 \\
& hybrid & 69.3\% & 79\% & 50\% & 379 \\
\midrule
\multirow{4}{*}{GPT-4o-mini}
& oracle & 86.7\% & 94\% & 72\% & 299 \\
& stm+sum+ltm & 84.7\% & 92\% & 70\% & 591 \\
& uniform & 81.3\% & 92\% & 60\% & 787 \\
& hybrid & 70.7\% & 80\% & 52\% & 379 \\
\bottomrule
\end{tabular}
\end{table}

\textbf{Finding 1: Store selection can improve accuracy and reduce tokens.}
Oracle routing outperforms uniform retrieval on both efficiency and answer quality. It achieves higher accuracy (86.7\% vs 81.3\%) while using 62\% fewer context tokens (299 vs 787). This result highlights that providing more context does not necessarily improve performance. When the router selects only the stores that are likely to contain the answer, the model receives a smaller but cleaner context, which can lead to more reliable extraction.

\textbf{Finding 2: Long context amplifies the penalty of over-retrieval.}
The difference between routing policies becomes more pronounced in the long-context setting. On long-context questions, oracle routing reaches 72\% accuracy compared with 60\% for uniform retrieval. When each store contributes roughly 1000 tokens, adding irrelevant stores significantly increases the amount of distracting text, making it harder for the model to identify the correct information.

\textbf{Finding 3: A strong fixed policy is competitive.}
A fixed routing policy such as STM+Sum+LTM approaches oracle-level accuracy while maintaining moderate cost. This suggests that even simple routing strategies can capture much of the benefit of selective retrieval when the memory architecture is well structured. Episodic retrieval, in contrast, is required only in a small subset of cases and can sometimes degrade performance by introducing unnecessary context.

\textbf{Finding 4: Heuristic routing is not yet sufficient.}
Although the hybrid heuristic achieves strong coverage on the synthetic routing benchmark, it performs less well on downstream QA tasks. This gap indicates that selecting the correct stores in principle does not always translate into better end-to-end performance. Downstream accuracy depends not only on store selection but also on the model’s ability to retrieve and use the relevant evidence within the provided context.

\subsection{Why Does Uniform Underperform?}

Uniform retrieval provides more information than oracle routing, yet it consistently yields lower accuracy. Two factors help explain this behavior.

\textbf{Needle in a haystack.}
With approximately 787 tokens in the prompt, the model must locate a small set of relevant facts embedded within a larger amount of unrelated text. This difficulty becomes more pronounced in long-context queries, where each additional store can contribute roughly 1000 tokens, further diluting the signal.

\textbf{Conflicting information.}
Different stores may contain inconsistent or outdated facts. For example, long-term memory may preserve earlier information that conflicts with updated user summaries. When all stores are retrieved together, the model must decide which source to trust, and it may occasionally select the wrong one.

\textbf{Example.}
Consider the query ``Who is my current manager?'' The Summary store contains the entry ``Manager: Jennifer Williams.'' The LTM store contains a previous session note: ``Before the reorg, user reported to Michael Torres. Now reports to Jennifer Williams.'' Under uniform retrieval, both passages appear in context, and GPT-4o-mini sometimes extracts ``Michael Torres'' from the more detailed historical passage. Under oracle routing, which retrieves only the Summary store, the model consistently returns ``Jennifer Williams.'' This example illustrates how additional context can sometimes mislead the model rather than improve performance.

\subsection{Understanding the Coverage-Accuracy Gap}

The hybrid heuristic achieves 94\% routing coverage (Section~\ref{sec:routing_eval}) but only 70\% QA accuracy. Several factors contribute to this difference.

\textbf{Coverage is not accuracy.}
Coverage measures whether the router included the stores that contain the required information. QA accuracy measures whether the model successfully identifies and uses that information in the retrieved context. Even when the correct store is present, extraction can fail if the context is long or contains distracting material.

\textbf{Missing stores remain decisive.}
When the hybrid heuristic fails to select a required store, which occurs in 6\% of queries, the question effectively becomes unanswerable under the retrieval setup. These cases directly contribute to the overall error rate.

\textbf{Over-retrieval can also hurt.}
The hybrid policy achieves 58\% exact match, meaning that 42\% of queries retrieve additional stores beyond those required. The resulting extra context can distract the model and reduce answer accuracy even when the correct stores are included.

Overall, the coverage-accuracy gap reflects two distinct failure modes. The first is routing error, where the heuristic misses a necessary store (12\% of QA errors). The second is extraction error, where the model fails to locate or correctly use the answer despite correct store selection (18\% of QA errors), often because over-retrieval introduces additional contextual noise.
\section{Analysis}
\label{sec:analysis}

Our results highlight two distinct phenomena. First, selecting the appropriate stores before retrieval can simultaneously reduce context tokens and improve downstream QA accuracy. Second, simple heuristic routing, although effective, does not fully close the performance gap relative to oracle routing. To better understand which signals drive routing quality, we analyze feature contributions and computational overhead.

\subsection{Feature Ablation}

We evaluate different feature groups on the synthetic routing benchmark to understand which signals are most useful for identifying the correct stores.

\begin{table}[H]
\centering
\caption{Feature ablation: routing coverage by feature type.}
\vspace{0.5em}
\begin{tabular}{lcc}
\toprule
Features & Coverage & $\Delta$ \\
\midrule
Linguistic (pronouns, tense) & 57\% & baseline \\
+ Semantic (quantity, temporal, multi-hop) & 90\% & +33\% \\
+ Embedding similarity & 94\% & +4\% \\
\bottomrule
\end{tabular}
\end{table}

\textbf{Semantic signals dominate.}
Adding semantic indicators such as quantity (``list all''), temporal references (``before''), and multi-hop reasoning cues (``compare'') increases coverage by 33 percentage points. These signals capture query patterns that simple linguistic features, such as pronouns or tense, often fail to identify. As a result, routing policies that rely only on shallow linguistic patterns tend to miss a substantial fraction of queries requiring historical or multi-store reasoning.

Embedding similarity provides an additional 4 percentage point improvement. Its main benefit appears on queries that do not contain clear surface cues but still have a semantic relationship to specific stores. In these cases, similarity scores help guide the router toward likely stores even when explicit rule-based triggers are absent.

\subsection{Computational Overhead}

Routing introduces minimal additional latency. Rule-based routing policies add less than 1\,ms of processing time, while embedding-based signals add approximately 5\,ms. Both overheads are small relative to typical LLM inference times, which range from 500 to 2000\,ms. 

Because routing reduces the amount of retrieved context, the resulting 62\% token reduction directly translates into lower inference cost. In practice, this reduction also decreases prompt processing time and can improve system responsiveness without requiring any modification to the underlying language model.
\section{Limitations}

\textbf{Synthetic labels.}
Ground-truth store labels are derived from query taxonomies rather than human annotation. This protocol allows controlled and reproducible evaluation of routing behavior, but it does not fully capture the variability present in real-world deployments. In practice, the necessity of a store may depend on how information is written, summarized, or updated over time. Human validation of store requirements would therefore provide a stronger assessment of routing accuracy in production settings.

\textbf{Heuristic router gap.}
The hybrid heuristic achieves high coverage on the synthetic routing benchmark but underperforms on downstream QA compared with oracle routing and strong fixed policies. This difference suggests that routing decisions interact closely with within-store retrieval quality and answer extraction. Closing the gap will likely require learned routing approaches that jointly optimize store selection and evidence retrieval rather than relying solely on rule-based heuristics.

\textbf{Two model families.}
Our evaluation includes GPT-3.5-turbo and GPT-4o-mini. While these models represent commonly used systems, testing additional architectures and training paradigms would strengthen claims about generalization. In particular, models with different context handling strategies or retrieval sensitivities may respond differently to routing policies.

\textbf{Full store retrieval.}
We concatenate full store contents instead of performing top-$k$ retrieval within each store. This design simplifies the analysis by isolating store-selection effects, but production systems often apply ranking or filtering within each memory store. The interaction between routing decisions and within-store retrieval strategies remains an important direction for future study.
\section{Conclusion}

We formalized memory store selection as a routing problem and evaluated it in two stages: first using synthetic labels to validate store-selection behavior, and then using real LLMs to measure downstream question answering performance. This two-stage evaluation separates the quality of routing decisions from the effects of retrieval and generation, allowing us to analyze the role of store selection more directly.

Our results show several consistent patterns. Selective store choice can improve both efficiency and accuracy, reducing context tokens by 62\% while increasing QA accuracy (86\% vs 81\%). Uniform retrieval, in contrast, can introduce unnecessary context that reduces performance, particularly in long-context settings where additional stores contribute large amounts of irrelevant text. Semantic signals substantially improve routing quality, increasing coverage by 33 percentage points compared with linguistic features alone. A simple fixed policy such as STM+Sum+LTM also provides a strong and deployable default when adaptive routing is not available.

We additionally introduced a decision-theoretic formulation of store routing that treats memory access as a cost-sensitive optimization problem. This framework explains the observed accuracy–efficiency tradeoffs and clarifies why routing decisions can influence answer quality even when retrieval and language models remain unchanged.

The remaining 16-point gap between heuristic routing (70\%) and oracle routing (86\%) indicates that further gains are possible. Closing this gap will likely require routing policies that are learned end-to-end and optimized directly for downstream QA outcomes rather than relying solely on hand-designed heuristics. We hope this work encourages further research on learned routing methods and highlights the importance of store selection as a core component of memory-augmented agent systems.
\section{Query Type to Store Mapping}
\label{sec:questions}

To evaluate routing behavior under controlled conditions, we assign each query type a set of stores that contain the information required to answer it. The mapping reflects how information is distributed across the memory architecture rather than how any particular retrieval system performs. For example, factual user attributes are stored in the Summary store, while historical comparisons typically require information from both Long-Term Memory and Episodic Memory. 

The goal of this mapping is to provide a consistent reference for measuring routing quality across policies. Because the mapping is derived from benchmark query taxonomies, it remains reproducible across datasets and does not depend on model outputs or manual labeling decisions. Table~\ref{sec:questions} lists the resulting query-type to store assignments used throughout our synthetic routing evaluation.

\begin{table}[H]
\centering
\caption{Ground-truth store labels by query type.}
\begin{tabular}{lcc}
\toprule
Query Type & Stores & Rationale \\
\midrule
single\_hop & Sum & Simple fact lookup \\
single\_session & STM & References current conversation \\
recent\_session & LTM & References past conversations \\
multi\_hop & Sum + LTM & Combines facts with history \\
memory\_capacity & LTM + Epi & Exhaustive recall (``list all'') \\
temporal & LTM + Epi & Historical comparison (``before X'') \\
knowledge\_update & Sum + LTM & Current vs historical state \\
\bottomrule
\end{tabular}
\end{table}

\section{Full Policy Comparison}
\label{app:full}

\begin{table}[H]
\centering
\caption{All 12 policies (GPT-4o-mini, 150 questions).}
\begin{tabular}{lcccc}
\toprule
Policy & Accuracy & Short & Long & Tokens \\
\midrule
oracle & 86.7\% & 94\% & 72\% & 299 \\
stm+sum+ltm & 84.7\% & 92\% & 70\% & 591 \\
uniform & 81.3\% & 92\% & 60\% & 787 \\
summary+ltm & 74.0\% & 82\% & 58\% & 406 \\
hybrid & 70.7\% & 80\% & 52\% & 379 \\
ltm & 49.3\% & 58\% & 32\% & 212 \\
ltm+episodic & 49.3\% & 57\% & 34\% & 408 \\
stm+summary & 45.3\% & 48\% & 40\% & 379 \\
summary & 30.7\% & 36\% & 20\% & 195 \\
episodic & 30.7\% & 43\% & 6\% & 196 \\
stm & 14.7\% & 14\% & 16\% & 184 \\
none & 14.0\% & 2\% & 38\% & 0 \\
\bottomrule
\end{tabular}
\end{table}
\section{Store-Access Cost Analysis}
\label{sec:storecost}

In addition to token cost, we examined the relative infrastructure cost associated with accessing different memory stores. In practical deployments, stores may reside on different storage tiers or require different retrieval pipelines, which can lead to varying access latency and compute overhead. To approximate these differences, we assign relative store-access costs: STM=1, Summary=1, LTM=3, and Episodic=5. These values represent normalized relative effort rather than exact system measurements.

Under this model, uniform retrieval incurs the highest cost (10), since it accesses all stores for every query. Oracle routing averages 3.9, reflecting its ability to select only the stores required for a given question. The hybrid heuristic reduces the average store-access cost by 29\% while maintaining 94\% coverage, demonstrating that routing policies can meaningfully reduce system overhead even when implemented with simple decision rules.

Although the main paper focuses on token cost because it directly influences LLM inference expense, the store-access analysis shows that similar efficiency gains appear at the infrastructure level. As routing policies become more adaptive, reductions in store-access overhead may translate into lower latency and improved scalability in production memory systems.
\bibliographystyle{iclr2026_conference}  
\bibliography{refs}
\appendix
\section{Memory Content Examples}
\label{app:memory}

To illustrate how information is distributed across stores, we present representative examples drawn from the synthetic memory construction used in our experiments. These examples demonstrate the semantic roles of each store rather than the exact evaluation instances.

\textbf{Short-Term Memory (STM).}
STM contains recent conversational context, such as scheduling or clarification requests:
\begin{quote}
``User just mentioned they have a meeting at 3pm today with the marketing team.'' \\
``User asked to schedule a follow-up call for tomorrow morning.''
\end{quote}
Queries such as ``What did I just say about today's meeting?'' require STM.

\textbf{Summary Store.}
The Summary store contains stable user facts and preferences:
\begin{quote}
``User Profile: Name is Alex Chen. Works at TechCorp as Senior Software Engineer.'' \\
``Contact: Phone number is 555-867-5309. Email is alex.chen@techcorp.com.'' \\
``Manager: Jennifer Williams.''
\end{quote}
Fact lookup queries such as ``Who is my manager?'' primarily depend on this store.

\textbf{Long-Term Memory (LTM).}
LTM stores summaries of past conversations:
\begin{quote}
``Session 9 (yesterday): Before the recent reorg, user's manager was Michael Torres. Now reports to Jennifer Williams.'' \\
``Session 10 (last week): User mentioned weight was 185 lbs before starting a diet.''
\end{quote}
Queries involving historical comparisons or past discussions typically require LTM.

\textbf{Episodic Memory.}
Episodic memory preserves raw conversation turns:
\begin{quote}
``Session 10, Turn 2: User said 'Back in January I was 185 pounds. With the diet I started, I'm hoping to get down to 170 by summer.' ''
\end{quote}
Queries requiring exact wording or timestamped references may depend on episodic retrieval.
\section{Code Availability}
\label{app:code}

All code used to generate synthetic routing labels, run routing-policy evaluations, 
and reproduce the LLM-based QA experiments is available in following repository:

\begin{center}
\url{https://github.com/krimler/memroute}
\end{center}

The repository includes scripts for synthetic memory generation, routing-policy 
evaluation, ablation studies, and end-to-end QA benchmarking. Instructions for 
reproducing the experimental results are provided in the project README. 
All experiments can be executed using the configuration files included in the repository.
\end{document}